\pdfoutput=1
\documentclass{article} % For LaTeX2e
\usepackage[utf8]{inputenc} % allow utf-8 input
\usepackage[T1]{fontenc}    % use 8-bit T1 fonts
\usepackage{hyperref}       % hyperlinks
\usepackage{url}            % simple URL typesetting
\usepackage{booktabs}       % professional-quality tables
\usepackage{amsfonts}       % blackboard math symbols
\usepackage{nicefrac}       % compact symbols for 1/2, etc.
\usepackage{microtype}      % microtypography
\usepackage{geometry}
\usepackage{authblk}
\usepackage{latexsym}

\usepackage{enumerate}
\usepackage[inline]{enumitem}
\usepackage{amsmath,amssymb}
\usepackage{amsfonts,dsfont}
\usepackage{nicefrac}
\usepackage{microtype}
\usepackage{mathtools}

\usepackage{caption}
\usepackage{subcaption}
\usepackage{wrapfig}
\usepackage{enumitem}
\usepackage{algorithm}
\usepackage[noend]{algorithmic}
\usepackage[normalem]{ulem}
\usepackage{amssymb}
\usepackage{multicol}
\usepackage{adjustbox}
\usepackage{multirow}
\usepackage{color}
\usepackage{xspace}
\usepackage{CJKutf8}

\PassOptionsToPackage{numbers}{natbib}
\usepackage{natbib}

\usepackage{mkolar_definitions}

\def\BibTeX{{\rm B\kern-.05em{\sc i\kern-.025em b}\kern-.08emT\kern-.1667em\lower.7ex\hbox{E}\kern-.125emX}}

\title{Optimal Transport-based Alignment of \\
	Learned Character Representations for String Similarity}

\author{Derek Tam$^1$, Nicholas Monath$^1$, Ari Kobren$^1$, Aaron Traylor$^2$, Rajarshi Das$^1$, Andrew McCallum$^1$ \\
$^1$College of Information and Computer Sciences, University of Massachusetts Amherst \\
$^2$Department of Computer Science, Brown University \\
\texttt{\{dptam,nmonath,akobren,rajarshi,mccallum\}@cs.umass.edu} \\
\texttt{aaron\_traylor@brown.edu}}

\date{}

\begin{document}
\maketitle

%auto-ignore
\newcommand{\alg}{\textsc{STANCE}\xspace}
\newcommand{\pstar}{\ensuremath{\mathcal{P}^\star}}
\newcommand{\phat}{\ensuremath{\widehat{\mathcal{P}}}}
\newcommand{\model}{\textsc{STANCE}\xspace}
\newcommand{\neural}{\textsc{LSTM-Dot}\xspace}
\newcommand{\minusot}{\textsc{Without-OT}\xspace}
\newcommand{\minuscnn}{\textsc{CNN-to-Linear}\xspace}
\newcommand{\minuslstm}{\textsc{LSTM-to-Binary}\xspace}

%auto-ignore
\begin{abstract}
  String similarity models are vital for record linkage, entity resolution, and
  search. In this work, we present \alg--a \emph{learned} model for
  computing the similarity of two strings. Our approach encodes the
  characters of each string, aligns the encodings using Sinkhorn
  Iteration (alignment is posed as an instance of optimal transport)
  and scores the alignment with a convolutional neural network. We
  evaluate \alg's ability to detect whether two strings \emph{can}
  refer to the same entity--a task we term \emph{alias detection}. We
  construct five new alias detection datasets (and make them publicly
  available). 
  We show that \alg (or one of its variants) outperforms both state-of-the-art and
  classic, parameter-free similarity models on four of the five
  datasets. 
  We also demonstrate \alg's ability to improve downstream
  tasks by applying it to an instance of cross-document coreference
  and show that it leads to a 2.8 point improvement in $B^3$ F1 over
  the previous state-of-the-art approach.
\end{abstract}

%auto-ignore
\section{Introduction}
String similarity models are crucial in record linkage, data
integration, search and entity resolution systems, in which they are
used to determine whether two strings refer to the same
\emph{entity}~\cite{bilenko2003adaptive,mccallum2005conditional,li2015robust}. In
the context of these systems, measuring string similarity is
complicated by a variety of factors including: the use of nicknames
(e.g., \texttt{Bill Clinton} instead of \texttt{William Clinton}),
token permutations (e.g., \texttt{US Navy} and \texttt{Naval Forces of
  the US}) and noise, among others. Many state-of-the-art systems
employ either classic similarity models, such as Levenshtein, longest
common subsequence, and Jaro-Winkler, or \emph{learned} models for
string
similarity~\cite{levin2012citation,li2015robust,ventura2015seeing,kim2016random,gan2017character}.

While classic and learned approaches can be effective, they both
have a number of shortcomings. First, the classic approaches have few
parameters making them inflexible and unlikely to succeed across
languages or across domains with unique characteristics (e.g. company
names, music album titles,
etc.)~\cite{needleman1970general,smith1981identification, winkler1999state,gionis1999similarity,bergroth2000survey,cohen2003comparison}. Classic
models also assume that each edit has equal cost, which is
unrealistic. For example, consider the names \texttt{Chun How} and
\texttt{Chun Hao}--which can refer to the same entity--and the names
\texttt{John A. Smith} and \texttt{John B. Smith}, which cannot.  Even
though the first pair differ by 2 edits and the second pair by 1,
transforming \texttt{ow} to \texttt{ao} in the first pair should cost
less than transforming \texttt{A} to \texttt{B} in the second. Learned
string similarity models address these problems by learning distinct
costs for various edits and have thus proven successful in a number of
domains~\cite{bilenko2003adaptive,mccallum2005conditional,gan2017character}.
Some learned string similarity models, such as the SVM \cite{bilenko2003adaptive} and CRF-based  \cite{mccallum2005conditional} approaches, use edit patterns akin to insertions/swaps/deletions, which may lead to strong inductive biases. 
For example, even when
costs are learned, two strings related by a token permutation--e.g.,
\texttt{Grace Hopper} and \texttt{Hopper, Grace}--are likely to have
high cost even though they clearly refer to the same entity.
\citet{gan2017character}, on the other hand, provide less
structure, encoding each string with a single vector embedding and measuring similarity between the embedded representations. 

In this paper, we present a learned string similarity model that is
flexible, captures sequential dependencies of characters,
and is readily able to learn a wide range of edit
patterns--such as token permutations. Our approach is comprised of three
components: the first encodes each character in both strings using a
recurrent neural network; the second softly aligns the two encoded
sequences by solving an instance of optimal transport; the third
scores the alignment with a convolutional neural network. Each
component is differentiable, allowing for end-to-end training. Our
model is called \alg--an acronym that stands for: \textbf{S}imilarity
of \textbf{T}ransport-\textbf{A}ligned \textbf{N}eural
\textbf{C}haracter \textbf{E}ncodings.

We evaluate \alg's ability to capture string similarity in a task we
term \emph{alias detection}. The input to alias detection is a query
\emph{mention} (i.e., a string) and a set of candidate mentions, and
the goal is to score query-candidate pairs that \emph{can} refer to
the same \emph{entity} higher than pairs that cannot. For example, an
accurate model scores the query \texttt{Philips} with candidates
\texttt{Philips Corporation} and \texttt{Katherine Philips} higher
than with \texttt{M. Phelps}. Alias detection differs from
both coreference and entity linking in that neither surrounding natural language context of the mention nor
external knowledge are available. A similar task is studied in recent
work~\cite{gan2017character}.

In experiments, we compare \alg to state-of-the-art and classic
models of string similarity in alias detection on 5 newly constructed
datasets--which we make publicly available.  Our results demonstrate
that \alg outperforms all other approaches on 4 out of 5 datasets in terms of Hits@1 and 3 out of 5 datasets in
terms of mean average precision. Of the two cases in which \alg is outperformed
by other methods in terms of mean average precision, one is by a variant of \alg in an ablation study.
We also demonstrate \alg's capacity for
supporting downstream tasks by using it in cross-document coreference
for the Twitter at the Grammy's dataset~\cite{dredze2016twitter}. 
Using \alg improves upon the state-of-the-art by 2.8 points of $B^3$ F1. Analyzing our trained model reveals \alg effectively learns
sequence-aware character similarities, filters noise with optimal
transport, and uses the CNN scoring component to detect unconventional
similarity-preserving edit patterns.

%auto-ignore
%auto-ignore
\begin{figure*}[t!]
  \centering
  \centerline{\includegraphics[width=0.9\textwidth]{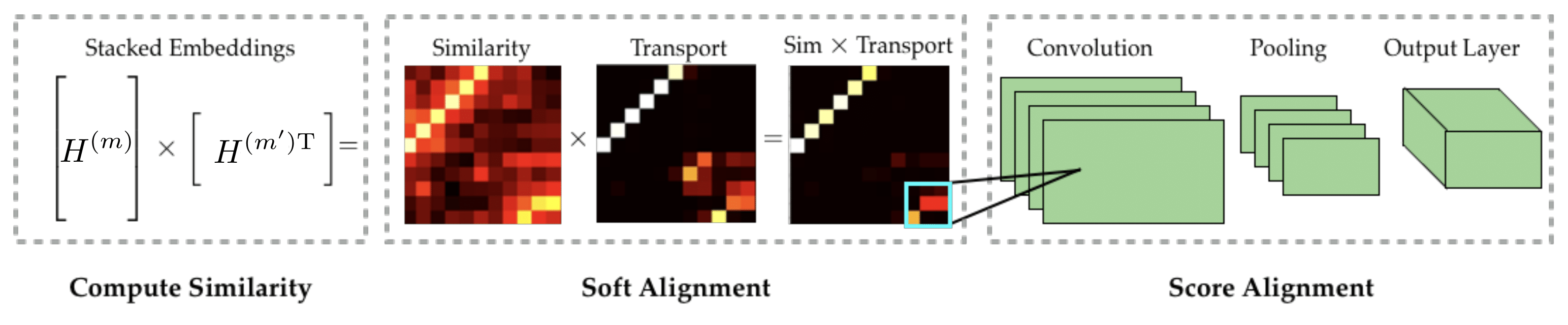}}
    \caption{\textbf{\alg Model architecture:} Character Similarities (\S \ref{subsec:sim}), soft alignment (\S \ref{subsec:align}), and scoring (\S \ref{subsec:score})}
  \label{fig:arch}
\end{figure*}
\section{\alg}
\label{sec:model}
Our goal is to learn a model, $f(\cdot, \cdot)$, that measures the
similarity between two strings--called \emph{mentions}. The model
should produce a high score when its inputs are \emph{aliases} of the
same entity, where a mention is an alias of an entity if it can be
used to refer to that entity. For example, the mentions \texttt{Barack
  H. Obama} and \texttt{Barry Obama} are both aliases of the entity
\texttt{wiki/Barack\_Obama}. Note that the alias relationship is not
transitive: both of the pairs \texttt{Obama}-\texttt{Barack Obama} and
\texttt{Obama}-\texttt{Michelle Obama} are aliases of the same entity,
but the pair \texttt{Barack Obama}-\texttt{Michelle Obama} are not.

In this section we describe our proposed model, \alg, which is
comprised of three stages: encoding both mentions and constructing a
corresponding similarity matrix, softly aligning the encoded mentions,
and scoring the alignment.

\subsection{Mention Encoding Similarity Matrix}
\label{subsec:sim}
A flexible string similarity model is sequence-aware, i.e., the cost of
each character transformation should depend on the surrounding
characters (e.g., transforming \texttt{Chun How} to \texttt{Chun Hao}
should have low cost).  To capture these sequential dependencies, \alg encodes each mention
using a bidirectional long short-term memory network
(LSTM)~\cite{hochreiter1997long,graves2005framewise}. In particular,
each character $c_i$ in a mention $m$ is represented by a
$d$-dimensional vector, $h_i$, where $h_i$ is the concatenation of the
hidden states corresponding to $c_i$ produced by running the LSTM in
both directions. The encoded representations of the characters are
stacked to form a matrix $H^{(m)} \in \mathbb{R}^{L\times d}$ where
$L$ (a hyperparameter) is the maximum string length considered by
\alg.

Given a query $m$ and candidate $m'$, \alg computes a \emph{similarity
  matrix} of their encodings via an inner product:
$S = H^{(m)} H^{(m')\mathrm{T}}$.  Each cell in the resultant matrix
represents a measure of the similarity between each pair of character
encodings from $m$ and $m'$.  Note that for a mention $q$ only the
first $|q|$ (i.e., length of the string $q$) rows of $H^{(q)}$ contain
non-zero values.

\subsection{Soft Alignment via Optimal Transport}
\label{subsec:align}
%auto-ignore
\begin{figure*}[t!]
  \centering
  \begin{subfigure}[t]{0.33\textwidth}
    \centerline{\includegraphics[height=0.75\textwidth]{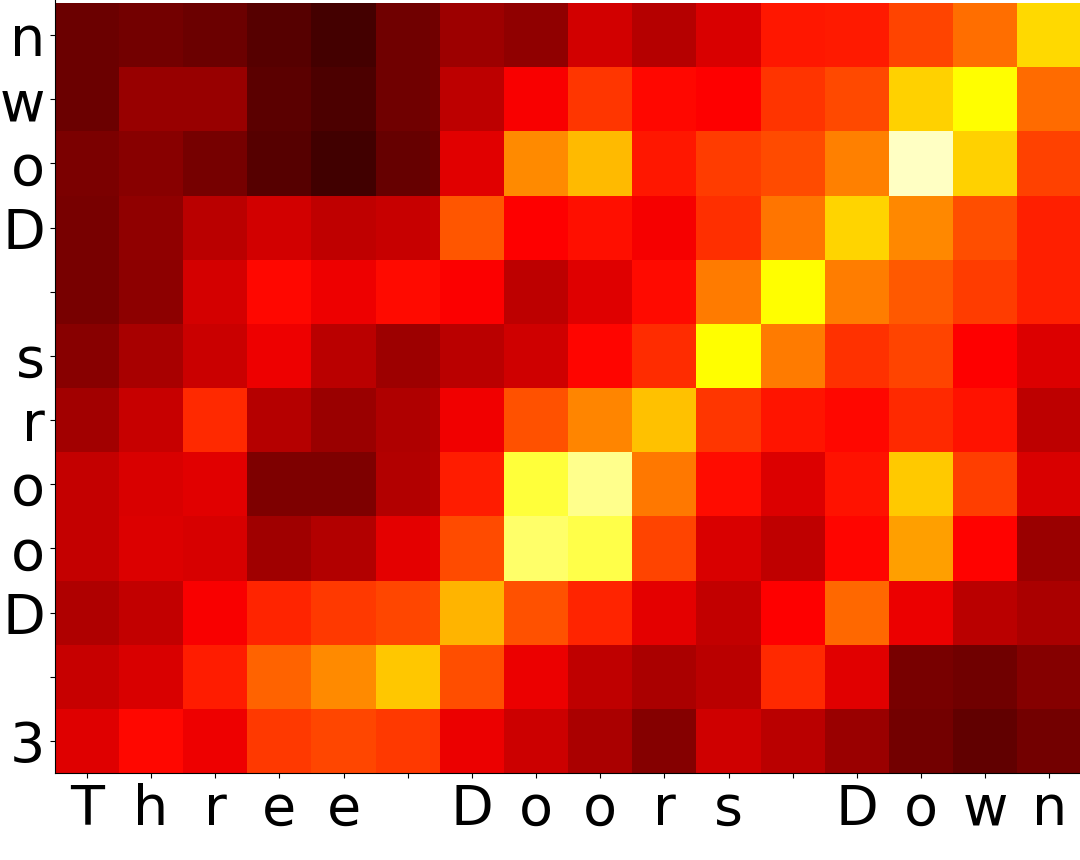}}
    \caption{Similarity Matrix}
    \label{fig:ot-sim}
  \end{subfigure}%
  \begin{subfigure}[t]{0.33\textwidth}
    \centerline{\includegraphics[height=0.75\textwidth]{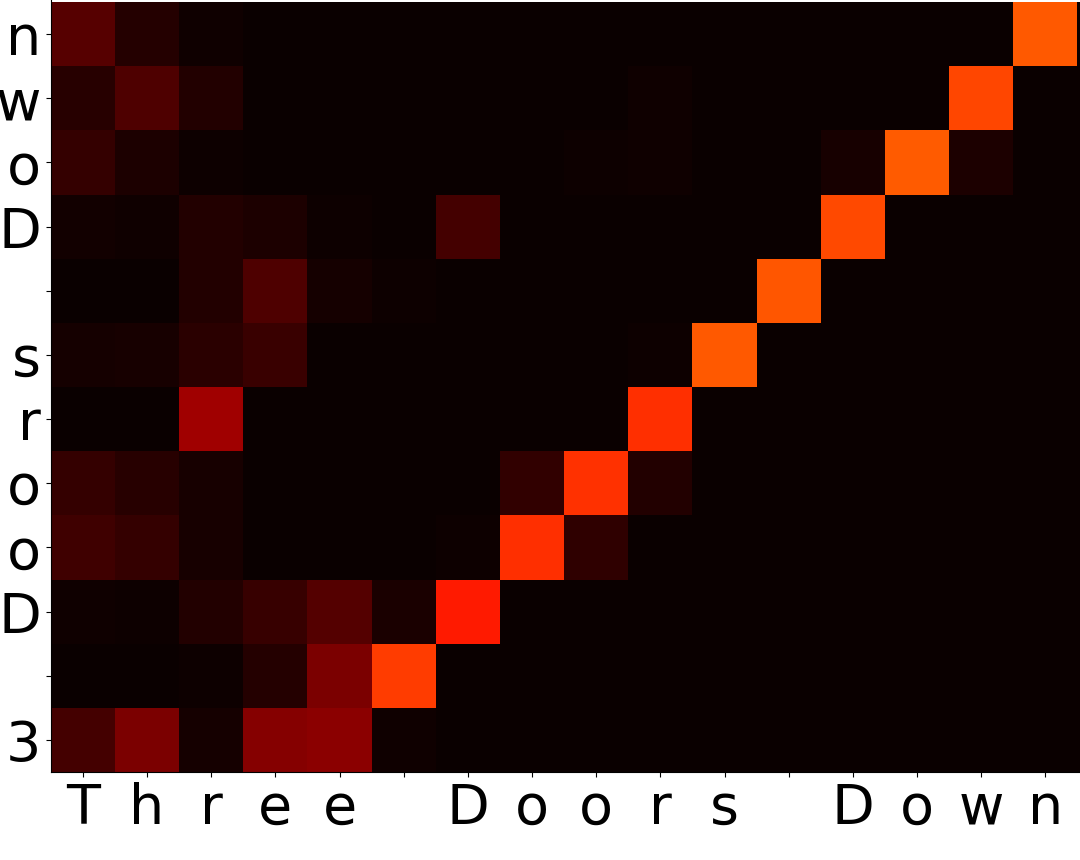}}
    \caption{Transport Matrix}
    \label{fig:ot-trans}
  \end{subfigure}%
  \begin{subfigure}[t]{0.33\textwidth}
    \centerline{\includegraphics[height=0.75\textwidth]{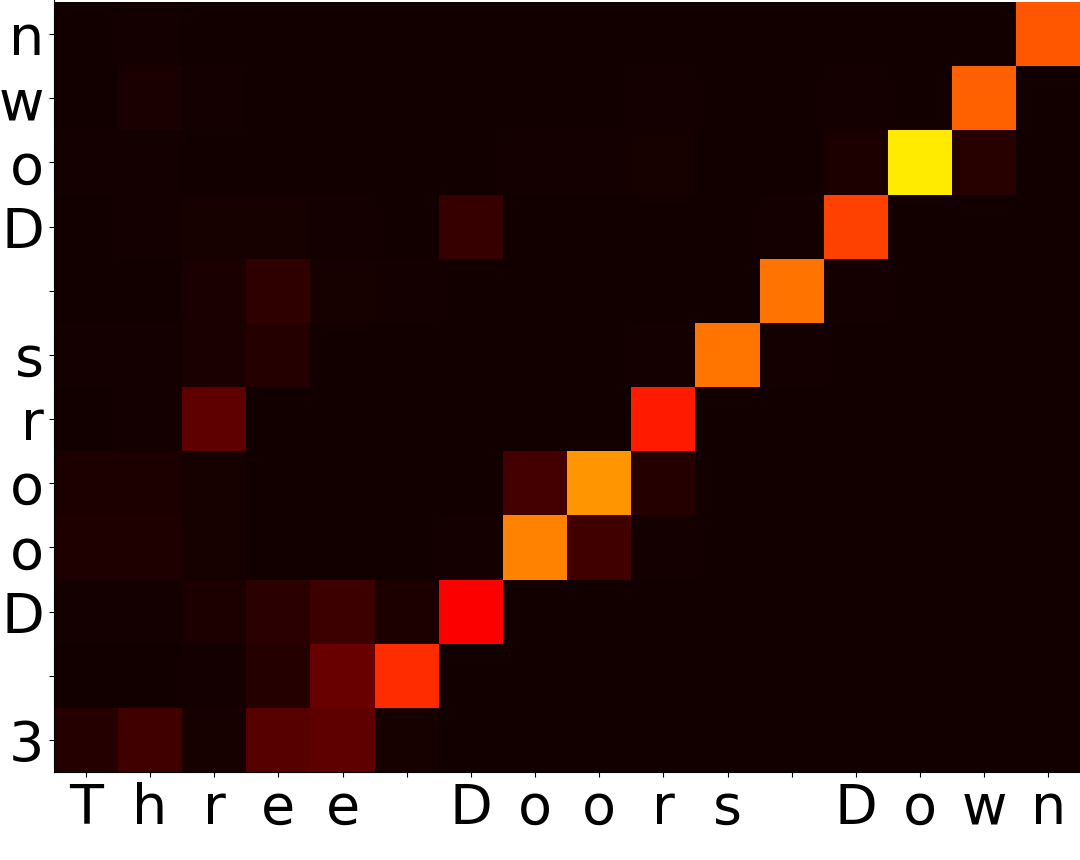}}
    \caption{Similarity $\times$ Transport}
    \label{fig:ot-simtrans}
  \end{subfigure}%
  \caption{\textbf{Three Heatmaps:} in all three heatmaps, brighter
    cells correspond to higher similarity. Figure \ref{fig:ot-sim}
    visualizes the character similarity matrix for two mentions:
    \texttt{Three Doors Down} and \texttt{3 Doors Down}.
    Figure \ref{fig:ot-trans} visualizes the transport matrix and Figure
    \ref{fig:ot-simtrans} visualizes the element-wise product of the
    similarity and transport matrices.  Many
    of the characters are highly similar.
    Multiplying by the transport matrix amplifies the alignment of the
    mentions while reducing noise, resulting in a clean alignment for the
    CNN scoring component.}
  \label{fig:heatmaps}
\end{figure*}

The next component of our model computes a soft alignment between the
characters of $m$ and $m'$. Aligning the mentions is posed as a
\emph{transport problem}, where the goal is to convert one mention
into another while minimizing cost.  In particular, we solve the
Kantorovich formulation of optimal transport (OT). In this
formulation, two probability measures, $p_1$ and $p_2$ are given in
addition to a cost matrix, $C$. This matrix defines the cost of moving
(or converting) each element in the support of $p_1$ to each element
in the support of $p_2$. The solution to OT is a matrix, $\hat{P}$,
called the \emph{transport plan}, which defines how to completely
convert $p_1$ into $p_2$.  A viable transport plan is required to be
non-negative and is also required to have marginals of $p_1$ and $p_2$
(i.e., if $\hat{P}$ is summed along the rows then $p_1$ is recovered
and if it is summed along the columns $p_2$ is recovered).  The goal is
to find the plan with minimal cost,
\vspace{-0.5cm}
\begin{align*}
P^\star &= \argmin_{P \in \Pcal}\sum_{i=0}^{|p_1|}\sum_{j=0}^{|p_2|}
C_{ij}P_{ij} \\
\Pcal &= \{ P \in \mathbb{R}_+^{L\times L} \ | \ P \mathbf{1}_L = p_1,\ P^T \mathbf{1}_L = p_2 \}
\end{align*}
where $|\cdot|$ is the number of elements in the support of the
corresponding distribution and $\Pcal$ is the set of valid
transportation plans. In this sense, a transportation plan can be
thought of as a soft alignment of the supports of $p_1$ and $p_2$
(i.e., an element in $p_1$ can be aligned fractionally to multiple
elements in $p_2$). A transportation plan can be computed efficiently
via Sinkhorn Iteration exploiting parallelism using GPUs (empirically it
has been shown to be quadratic in $L$) \cite{cuturi2013sinkhorn}.
The transport plan is defined as $P = \text{diag}(\pmb{u}) K \text{diag}(\pmb{v})$
where $K := e^{- \lambda C}$, $\pmb{u}$ and $\pmb{v}$ are found using
the iterative algorithm, $\lambda$ is the entropic regularizer, 
and \text{diag}$(\cdot)$ gives a matrix with its input argument as
 the diagonal \cite{cuturi2013sinkhorn}. We specifically use
the regularized objective that has been shown to be effective for
training \cite{cuturi2013sinkhorn,genevay2018learning}.

Optimal transport has been effectively used in several natural language-based 
applications
such as computing the similarity between two documents as 
the transport cost \cite{kusner2015word,huang2016supervised}, 
in measuring distances between point cloud-based representations of words \cite{frogner2018learning}, and
learning correspondences between word embedding spaces across domains/languages
\cite{alvarez2018gromov,pmlr-v89-alvarez-melis19a}.

In our case, $p_1$ represents the mention $m$ and $p_2$ represents
$m'$. The distribution $p_1$ is defined as a point cloud consisting of
the character embeddings computed by the LSTM applied to $m$, i.e.,
$H^{(m)}$. Formally, it is a set of evenly weighted Dirac Delta
functions in $\mathbb{R}^d$ where $d$ is the embedding dimensionality
of the character representations. The distribution $p_2$ is defined
similarly for $m'$. The cost of transporting a character, $c_i$ of $m$
to a character $c_j$ of $m'$ has cost, $C_{i,j} = S_{\max} - S_{i,j}$
where $S_{\max} = \max_{i',j'} S_{i',j'}$ and $S_{i,j}$ is the inner
product of $h_i$ and $h_j$. The resulting transport plan is
multiplied by the similarity matrix (Section \ref{subsec:sim}) and
subsequently fed as input to the next component of our model (Section
\ref{subsec:score}). Despite being a soft alignment, this step helps
mitigate spurious errors by reducing the similarity of characters
pairs that are not aligned.  

\subsection{Alignment Score}
\label{subsec:score}
The transport plan, $\hat{P} \in \mathbb{R}_+^{L \times L}$ describes
how the characters in $m$ are softly aligned to the characters in
$m'$.  We compute the element-wise product of the similarity matrix,
$S$, and the transport plan: $S' = S \circ \hat{P}$. Cells containing
high values in $S'$ correspond to similar character pairs from $m$ and
$m'$ that are also well-aligned.

Note the distinction between
this alignment and the way in which the transport cost can be
used as distance measure. The alignment is used as a re-weighting 
of the similarity matrix. In this way, the transport plan is closely related to attention-based
models \cite{bahdanau2015neural,parikh2016decomposable,vaswani2017attention,kim2017structured}.

Finally, we employ a two dimensional convolutional neural network (CNN) to score
$S'$~\cite{lecun1998gradient}. With access to the full matrix $S'$,
the CNN is able to detect multiple, aligned, character subsequences
from $m$ and $m'$ that are highly similar. By combining evidence from
multiple--potentially non-continguous-- aligned character
subsequences, the CNN detects long-range similarity-preserving edit
patterns. This is crucial, for example, in computing a high score for
the pair \texttt{Obama, Barack} and \texttt{Barack Obama}. 

The architecture of the alignment-scoring CNN is a three layer network
with filters of fixed size. A linear model is used to score the
final output of the CNN.  See Figure \ref{fig:arch} for a visual
representation of the \alg architecture.

\paragraph{Training}
We train on mention triples, $(q,p,n)$, where there exists an entity
for which $q$ and $p$ are both aliases (i.e., $(q,p)$ is a positive
example), and there does not exist an entity for which both $q$ and
$n$ are aliases (i.e., a negative example).  We use the Bayesian Personalized Ranking objective \cite{rendle2009bpr}:
$\sigma(f(q,p) - f(q,n))$.

%auto-ignore
\section{Alias Detection}
\label{sec:datasets}
%auto-ignore
%\begin{figure*}
%	\centering
%	\includegraphics[width=0.85\textwidth]{negative_examples.pdf}
%	\caption{Illustration of true positive aliases and the five types of true
%	negatives used in evaluation and described in the text. The figure depicts
%	the source knowledge base with mentions as ovals, entities as
%	squares, and the query in a red oval. Links indicate that an entity is
%	referred to by that mention.}
%	\label{fig:graph}
%\end{figure*}
\begin{figure}
	\centering
	\includegraphics[width=0.99\textwidth]{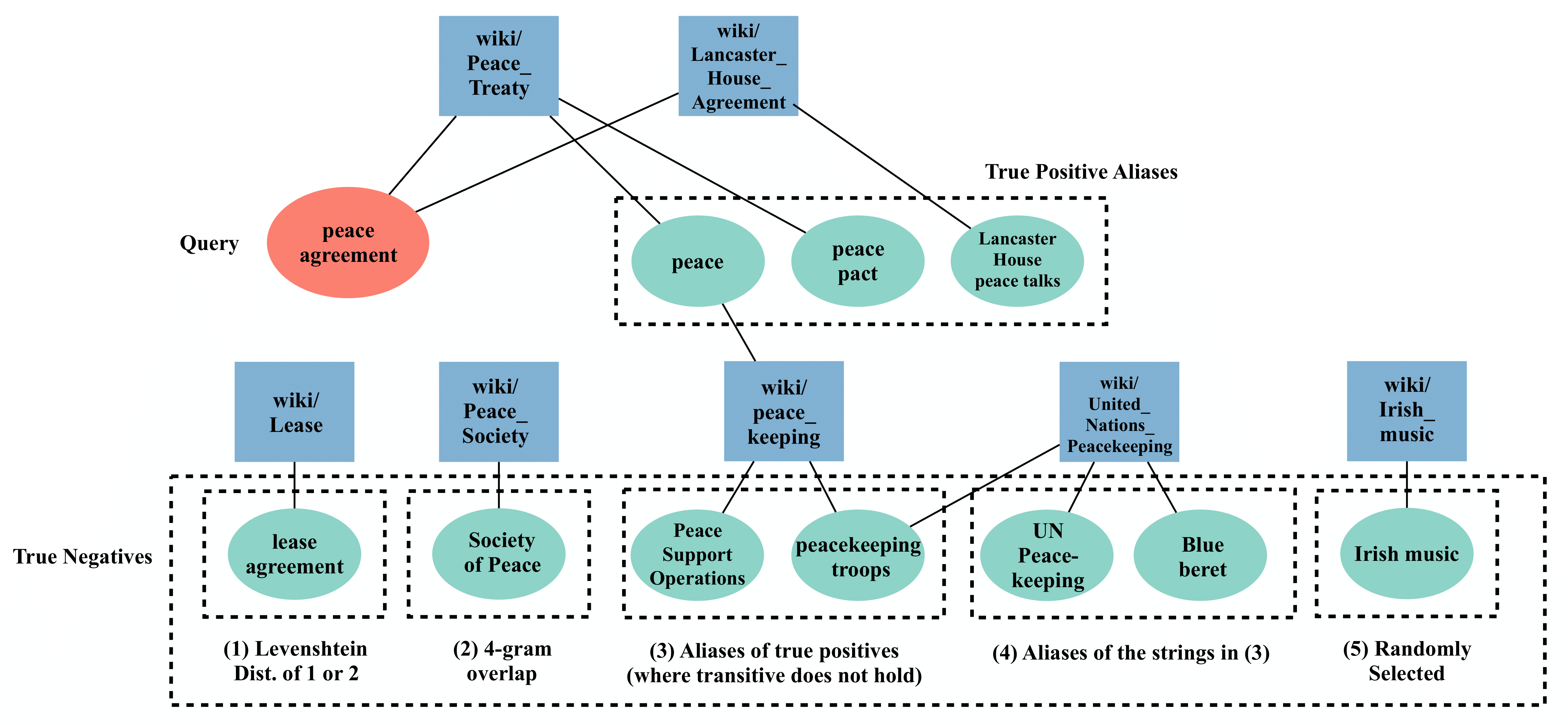}
	\caption{\textbf{True positive and negative aliases}. A depiction of
		the source KB with mentions as ovals, entities as
		squares, and the query in a red oval. Links indicate that an entity is
		referred to by that mention.}
	\label{fig:graph}
\end{figure}
String similarity is a crucial piece of data integration, search and
entity resolution systems, yet there are few large-scale
datasets for training and evaluating domain-specific string similarity
models. Unlike in coreference resolution, a high quality model should
return high scores for mention pairs in which both strings are aliases
of (i.e., \emph{can} refer to) the same entity. For example, the
mention \texttt{Clinton} should exhibit high score with both
\texttt{B. Clinton} and \texttt{H. Clinton}.

We construct five datasets for training and evaluating string
similarity models derived from four large-scale public knowledge bases, which
encompass a diverse range of entity types. The five datasets are
summarized below:
\begin{enumerate}[noitemsep,nolistsep]
\item \textbf{Wikipedia (W)} -- We consider pages in Wikipedia to be
  entities. For each entity, we extract spans of text hyperlinked to
  that entity's page and use these as aliases.\footnote{We used a xml dump of Wikipedia from 2016-03-05. We restrict the entities and hyperlinked spans to come from non-talk, non-list Wikipedia pages.}
\item \textbf{Wikipedia-People (WP)} -- The Wikipedia
dataset restricted to entities with type \texttt{person} in
Freebase ~\cite{bollacker2008freebase}.
\item \textbf{Patent Assignee (A)} --
Aliases of assignees (mostly organizations, some persons) found by combining entity information\footnote{\url{sites.google.com/site/patentdataproject/Home/downloads}} with non-disambiguated assignees in patents\footnote{\url{www.patentsview.org/}}.
\item \textbf{Music Artist (M)} -- MusicBrainz
  \cite{swartz2002musicbrainz} contains alternative names for music artists.
\item \textbf{Diseases (D)} -- The Comparative Toxicogenomics Database
\cite{davis2014comparative} stores alternative names for disease entities.
\end{enumerate}
%auto-ignore
\begin{table*}[t]
	\footnotesize
	\centering
	\resizebox{\columnwidth}{!}{
	\begin{tabular}{c c c c c c}
        \hline
		Data & Unique Strings & Entity Count &
                                                                  Avg. Num.
                                                                  of
                                                                  Mentions/Ent
          & Avg. TP/Ent (Dev) & Avg. TP/Ent (Test)\\
		\hline
		W & 9.32 $\times$ $10^6$& 4.64 $\times$ $10^6$ & 2.54 $\pm$ 4.65 & 125.01 $\pm$ 356.45 &  80.31 $\pm$ 317.42 \\
		WP & 1.88 $\times$ $10^6$ & 1.16 $\times$ $10^6$& 1.83  $\pm$ 2.06 & 9.82 $\pm$ 23.71 & 10.53 $\pm$ 43.35\\
		A & 3.30 $\times$ $10^5$ & 2.27 $\times$ $10^5$&1.501 $\pm$ 2.64 & 30.76  $\pm$ 63.46 & 11.42 $\pm$ 25.02  \\
		M & 1.83 $\times$ $10^6$ & 1.16 $\times$ $10^6$ & 1.694 $\pm$ 3.23 & 5.08 $\pm$ 13.63 & 9.20 $\pm$ 136.28\\
		D & 7.69 $\times$ $10^4$& 1.19 $\times$ $10^4$ & 6.67 $\pm$ 9.10 & 7.21 $\pm$ 10.60 & 7.46 $\pm$ 10.72 \\
	    \hline
    \end{tabular}
    }
    \caption{Qualities of the 5 created datasets. True positive are
      correct entity aliases included in the dev or test set.}
  	\label{tbl:datasets}
\end{table*}

For each dataset, entities are divided into training, development, and
testing sets, such that each entity appears in \emph{only one
  set}. This partitioning scheme is meant to ensure that performant
models capture a general notion of similarity, rather than learning to
recognize the aliases of particular entities.  Dataset statistics can be found in Table~\ref{tbl:datasets}.

Most mention-pairs selected uniformly at random are not aliases of the
same entity. A model trained on such pairs may learn to always predict
``Non-alias.'' 
 To avoid learning such degenerate models and to avoid
test sets for which degenerate models are performant, we carefully
construct the training, development and test sets by including a mix
of positive and negative examples and by generating negative examples
designed to be difficult and practical. We use a mixture of the
following five heuristics to generate negative examples:
\begin{enumerate}[noitemsep,nolistsep]
\item \textbf{Small Edit Distance} -- mentions with Levenshtein
  distance of 1 or 2 from the query;
\item \textbf{Character Overlap} -- mentions that share a
4-gram word prefix or suffix with the query;
\item \textbf{4-Hop Aliases} -- first, construct a bipartite graph of
  mentions and entities where an edge between a mention and an entity
  denotes that the mention is an alias of the entity. Then, sample a
  mention that is not an alias of an entity for which the query is
  also an alias, and whose shortest path to the query requires 4 hops
  in the graph. Note that all mentions 2 hops from the query are
  aliases of an entity for which the query is also an alias.
\item \textbf{6-Hop Aliases} -- sample a mention whose shortest path
  to the query in the bipartite mention-entity graph is 6 hops.
  \begingroup\interlinepenalty=10000
\item \textbf{Random} -- randomly sample mentions that are not aliases
  of the entity for which the query is also an alias. We do this by
  first sampling an entity and then sampling an alias of that entity
  uniformly at random.
\end{enumerate}

\endgroup

In all cases, we sample such that entities that appear more frequently
in the corpus and entities that have a larger number of aliases are
more likely to be sampled (intuitively, these entities are more
relevant and more challenging). For the Wikipedia-based datasets, we
sample entities proportionally to the number of hyperlink spans
linking to the entity. For the Assignee dataset, we estimate entity
frequency by the number of patents held by the entity. For the Music
Artist dataset, entity frequency is estimated by the number of entity
occurrences in the Last-FM-1k dataset
\cite{lastfm,Celma:Springer2010}. For the disease dataset, we do not
have frequency information and so sampling is performed uniformly
at random.  For each dataset, 300 queries are selected for use in the
development set and 4000 queries for use in the test set. Each query
is paired with up to 1000 negative examples of each type mentioned
above.  For training, we also construct datasets using the
approaches above for creating negative examples.

Figure \ref{fig:graph} illustrates how negative (and
positive) examples are generated for the query \texttt{peace agreement} (which is
used to refer to the entities \texttt{wiki/Peace\_Treaty} and \
\texttt{wiki/Lancaster\_House} \texttt{\_Agreement}). 4-Hop
(negative) aliases include \texttt{Peace Support Operations} and
\texttt{peacekeeping troops} and 6-Hop (negative) examples include
\texttt{UN Peacekeeping} and \texttt{Blue beret}. Note that for each
type of negative example, any mention that is a true positive alias of
the query is excluded from being a negative example, even if it
satisfies one of the above heuristics.

\section{Experiments}
\label{sec:exp}
%auto-ignore
\begin{table*}[t]
	\footnotesize
	\centering
	\resizebox{\columnwidth}{!}{
	\begin{tabular}{c c c c c c c c c c c c c c c}
                 & Ours           && \multicolumn{8}{c}{Alias Detection}  && \multicolumn{3}{c}{Ablation}            \\
         \cline{2-2}\cline{4-11}\cline{13-15}\\
        Data  & \model         && Lev    & JW     & LCS    & Sdx   & CRF   & LSTM  & DCM & LDTW && -CNN &  -LSTM & -OT      \\
        \hline
        W     & \textbf{.416} && .238  & .297  & .332  & .294 & .299 & .230 & .288 & .362 && .208     & .287       & .340         \\
        WP  & \textbf{.594} && .246  & .283  & .397  & .308 & .515 & .328 & .352 & .413 && .234     & .411       & .538         \\
        A & .906          && .720  & .850  & .622  & .733 & .780 & .790 &.782  & .903 && .797     & .838       & \textbf{.910}\\
        M   & \textbf{.597} && .296  & .328  & .293  & .354 & .319 & .399 &.509& .396  && .250     & .403       & .475         \\
        D  & .417 && .206  & .244  & .191  & .259 & .162 & .247  &  \textbf{.437} & .347 && .230     & .252       & .360         \\
        \hline
	\end{tabular}
	}
	\caption{Mean Average Precision (MAP).}
	\label{tbl:map}
\end{table*}
%auto-ignore
\begin{table*}[t]
	\footnotesize
	\centering
	\setlength{\tabcolsep}{0.75\tabcolsep}
	\begin{tabular}{c c c c c c c c c c c c c c c c}
                 &         & Ours           && \multicolumn{8}{c}{Alias Detection}     && \multicolumn{3}{c}{Ablation}               \\
        \cline{3-3}\cline{5-12}\cline{14-16}\\
        Data  & K  & \model         && Lev   & JW    & LCS   & Sdx   & CRF   & LSTM & DCM & LDTW && -CNN      & -LSTM      & -OT \\
        \hline
	   & 1  & \textbf{.698} && .553 & .630 & .569 & .545 & .599 & .436 & .610 & .570 && .358     & .509      & .586  \\
	W  & 10 & \textbf{.599} && .380 & .471 & .450 & .381 & .464 & .383  & .440 & .525 && .355     & .444      & .515 \\
	   & 50 & \textbf{.604} && .373 & .488 & .441 & .366 & .474 & .448 &.431  & .556 && .446     & .507      & .556\\
        \hline
	   & 1  & \textbf{.744} && .434 & .506 & .570 & .422 & .648 & .421 & .528 & .456 && .300     & .550      & .680 \\
	WP & 10 & \textbf{.708} && .397 & .397 & .475 & .323 & .646 & .469 & .459  & .573 && .357     & .544      & .665\\
	   & 50 & \textbf{.766} && .417 & .488 & .517 & .370 & .716 & .745 &.546 & .729 && .547     & .672      & .745  \\
        \hline
	   & 1  & \textbf{.942} && .850 & .920 & .726 & .808 & .867 & .863 & .881& .926  && .821     & .870      & .932 \\
	A  & 10 & .932          && .805 & .896 & .738 & .746 & .840 & .870  &.841 & .947 && .879     & .904      & \textbf{.950}  \\
	   & 50 & .966          && .847 & .930 & .817 & .789 & .896 & .927  &.883 & \textbf{.970} && .940     & .946      & \textbf{.970} \\
	\hline
	   & 1  & \textbf{.698} && .442 & .475 & .417 & .382 & .465 & .460 &.614 & .406  && .251     & .483      & .562 \\
	M  & 10 & \textbf{.690} && .369 & .386 & .398 & .328 & .371 & .538  &.623 & .532 && .388     & .525      & .581 \\
	   & 50 & \textbf{.806} && .448 & .506 & .502 & .430 & .452 & .707 &.746 & .716  && .595     & .682      & .743 \\
	\hline
	   & 1  & .589 && .514 & .517 & .458 & .451 & .410 & .449 & \textbf{.630} & .508 && .314     & .381      & .505 \\
	D  & 10 & \textbf{.521} && .266 & .300 & .285 & .260 & .232 & .329 &.499  & .455 && .334     & .349      & .475 \\
	   & 50 & \textbf{.638} && .305 & .395 & .371 & .324 & .316 & .470 & .571 & .600  && .497     & .511      & .604 \\
        \hline
	\end{tabular}
	\caption{Hits at K.}
	\label{tbl:hits}
\end{table*}

We evaluate \alg directly via alias detection and also indirectly via
cross document coreference.  We also conduct an ablation study in
order to understand the contribution of each of \alg's three
components to its overall performance.

\subsection{Alias Detection}
In the first experiment, we compare \alg with both classic and learned
similarity models in alias detection. Specifically, we compare \alg to
following approaches:

\begin{itemize}[noitemsep,nolistsep]
\item \textbf{Deep Conflation Model (DCM)} -- state of the art model
  that encodes each string using a 1-dimensional CNN applied to
  character n-grams and computes cosine
  similarity~\cite{gan2017character}.  We use the available code
  \footnote{\url{github.com/zhegan27/Deep_Conflation_Model}}.
\item \textbf{Learned Dynamic Time Warping (LDTW)} -- encode mentions
  using a bidirectional LSTM and compute similarity via dynamic time
  warping (DTW). We note equivalence between LDTW and weighted finite
  state transducers where the transducer topology is the edit distance
  (insert, delete, swap) program. Parameters are learned such that DTW
  distance is meaningful~\cite{cuturi2017soft}.
\item \textbf{LSTM} -- represent each mention using the final hidden
  state of a bidirectional LSTM. Similarity is the dot product of
  mention representations (i.e. $S_{|m||m'|}$).
\item \textbf{Classic Approaches} -- Levenshtein
  Distance (Lev), Jaro-Winkler distance (JW), Longest Common
  Subsequence (LCS).
\item \textbf{Phonetic Relaxation (Sdx)} -- transform mentions using the
  Soundex phonetic mapping and then compute Levenshtein.
\item \textbf{CRF} -- implementation \footnote{\url{github.com/dirko/pyhacrf}} of
  the model defined in \cite{mccallum2005conditional}.
\end{itemize}

Given a query mention, $q$, and a set of candidate mentions, we use
each model to rank candidates by similarity to
$q$. We compute the mean average precision (MAP) and hits at
$k=\{1, 10, 50\}$ of the ranking with respect to a set of ground truth
labeled aliases. We report MAP and hits at $k$ averaged over all test
queries. The set of candidates for query $q$ include all corresponding
positive and negative examples from the test set (Section
\ref{sec:datasets}).

For models with hyperparameters, we tune the hyperparameters on the
dev set using a grid search over:
embedding dimension, learning rate, hidden state dimension, and number of filters (for the
CNN).  All models were implemented in PyTorch, utilizing SinkhornAutoDiff \footnote{\url{github.com/gpeyre/SinkhornAutoDiff}}, and optimized with
Adam~\cite{kingmaleiba2015adam}.  Our \mbox{implementation} is publicly available~\footnote{\url{github.com/iesl/stance}}.

\subsection{Ablation Study}
\label{subsec:ablation}
Our second experiment is designed to reveal the purpose of each of
\alg's components. To do so, we compare variants of \alg with
components removed and/or modified. Specifically, we compare the
following variants:

\begin{itemize}[noitemsep,nolistsep]
\item \textbf{\minusot (-OT)} -- \alg with LSTM encodings and CNN scoring
but without optimal transport-based alignment.
\item \textbf{\minuscnn (-CNN)} -- \alg with the CNN scoring model replaced
by a linear scoring model. Again, the optimal transport-based alignment is removed.
\item \textbf{\minuslstm (-LSTM)} -- A binary similarity matrix
  ($S_{ij} = \mathbb{I}[m_i=m'_j]$) and CNN scoring model, designed to
  assess the importance of the initial mention encodings. Once more, the optimal transport-based alignment is removed.
\end{itemize}

We evaluate each model variant using MAP and hits at $k$ on the 5
datasets as in the first experiment.
Results can be found in Table \ref{tbl:map} and Table \ref{tbl:hits},
respectively. We note that these ablations are equivalent to the models
proposed by \citet{traylor2017}.

\subsection{Results and Analysis}
\label{subsec:analysis}
Table \ref{tbl:map} and Table \ref{tbl:hits} contain the MAP and hits
at $k$ (respectively) for each method and dataset (for alias detection
and ablation experiments).
The results reveal that with the exception of the disease dataset,
\alg (or one of its variants) performs best in terms of both
metrics. The results suggest that the optimal transport and CNN-based
alignment scoring components of \alg lead to a more robust model of
similarity than inner-product based models, like \textbf{LSTM} and
\textbf{DCM}. We hypothesize that using n-grams as opposed to
individual characters embeddings is advantageous on the disease
dataset, leading to \textbf{DCM}'s top performance. Surprisingly, -OT
is best on the assignee dataset. We hypothesize that this is due to
many corporate acronyms.

To better understand \alg's performance and improvement over the
baseline methods we provide analysis of particular examples
highlighting two advantages of the model: it leverages optimal
transport for noise reduction, and it uses its CNN-based scoring
function to learn non-standard similarity-preserving string edit
patterns that would be difficult to learn with classic edit operations
(i.e., insert, delete and substitute).

\begin{figure}[h]
  \begin{subfigure}[t]{0.49\textwidth}
    \centerline{\includegraphics[width=0.95\textwidth]{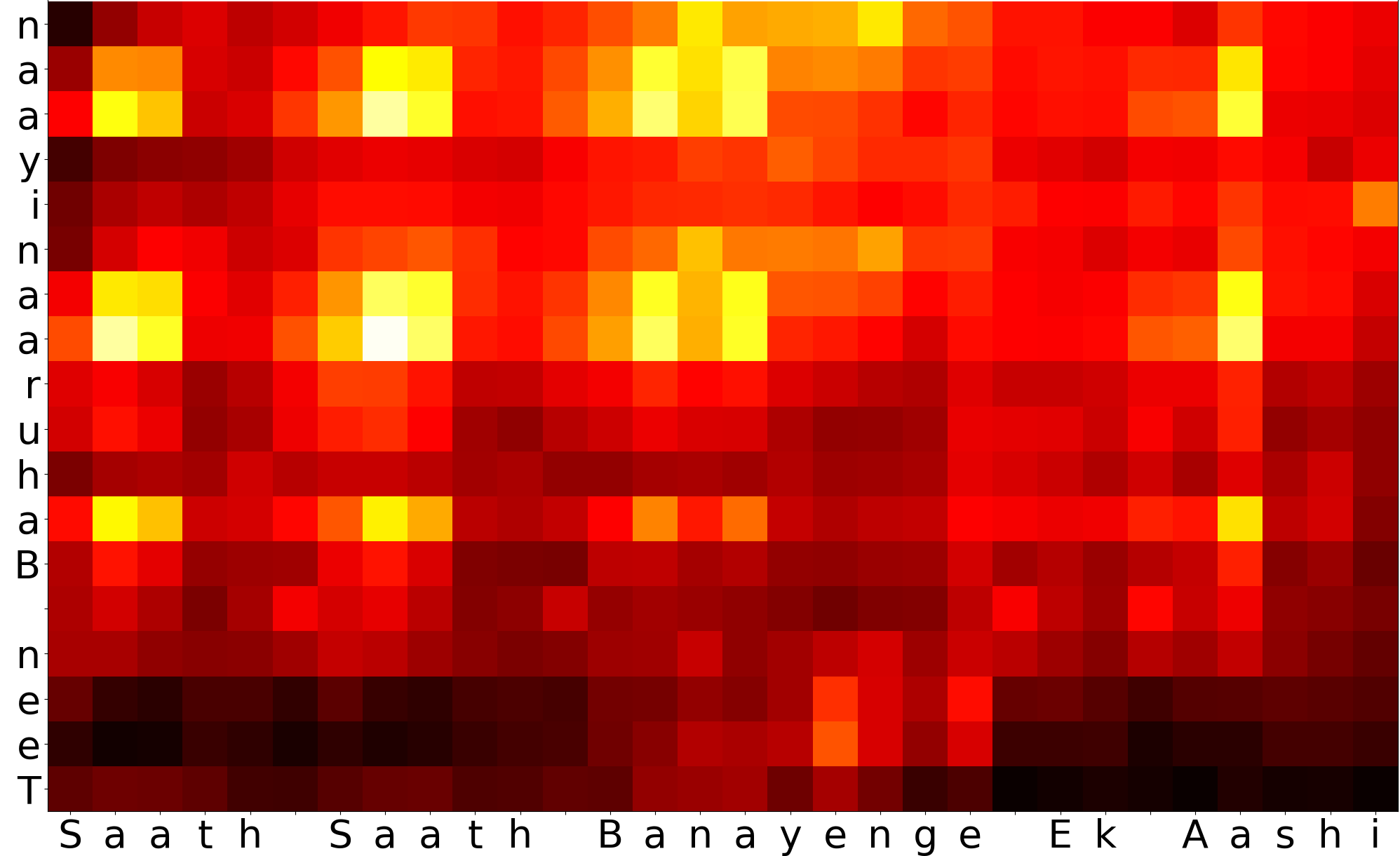}}
    \caption{Similarity Matrix.}
    \label{fig:flex}
  \end{subfigure}
  \begin{subfigure}[t]{0.49\textwidth}
    \centerline{\includegraphics[width=0.95\textwidth]{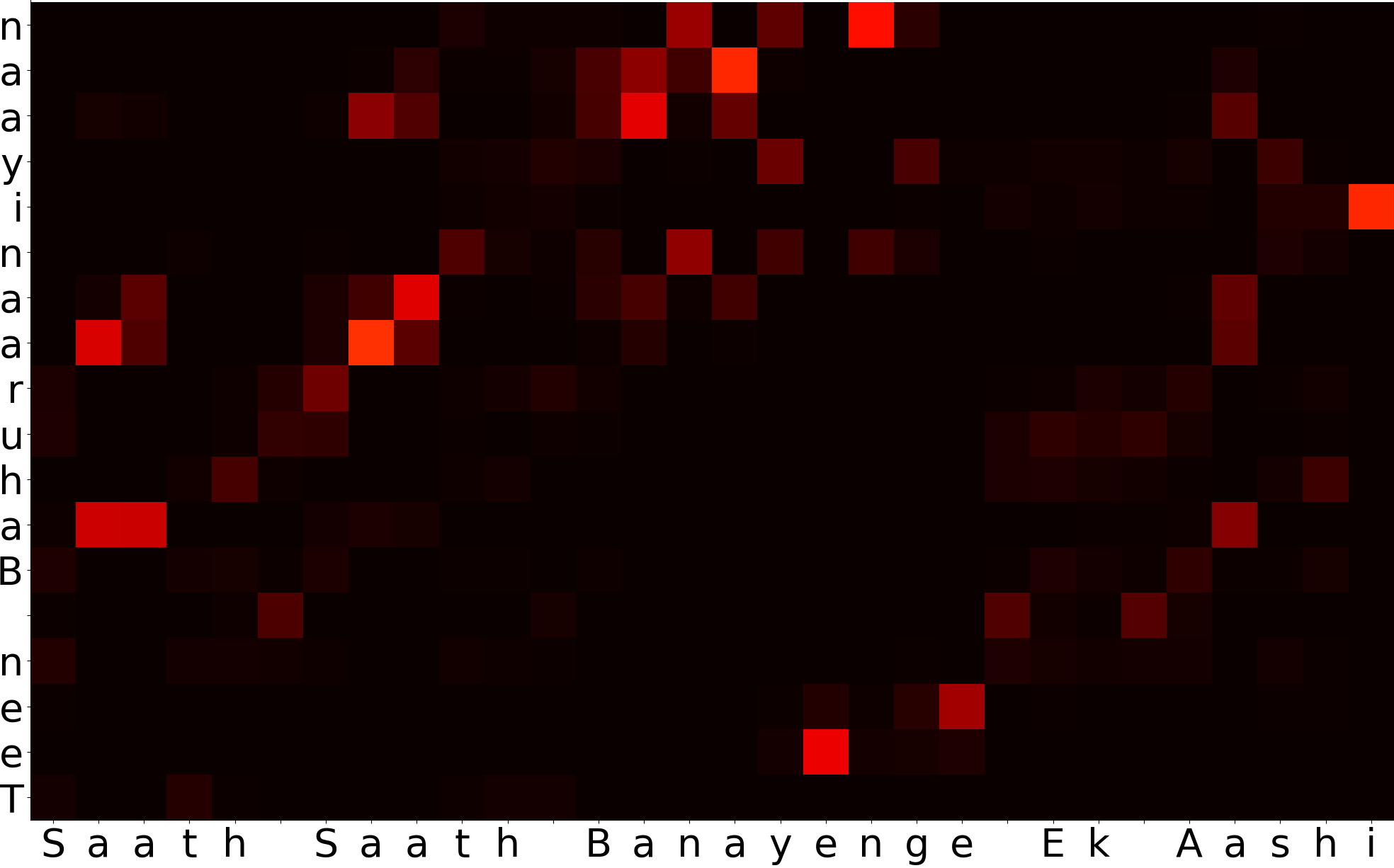}}
    \caption{Noise Filtered}
    \label{fig:noise}
  \end{subfigure}
  \caption{\textbf{Noise Filtering:} OT effectively
    reduces noise in the similarity matrix even when many character
    n-grams are common to both mentions (Teen Bahuraaniyaan / Saath Saath Banayenge Ek Aashi).}
\end{figure}

\paragraph{Noise Reduction.} Since the model leverages distributed
representations for characters, it often discovers many similarities between
the characters in two mentions.  For example, Figure \ref{fig:flex}
shows two strings that are not aliases of the same entity. Despite
this, there are many regions of high similarity due to multiple
instances of the character bigrams \texttt{aa}, \texttt{an} and
\texttt{en} in both mentions.  In experiments, we find that this leads
the -OT model astray.  However, \alg's optimal transport
component constructs a transport plan that contains little alignment
between the characters in the mentions as seen in Figure
\ref{fig:noise}, which displays the product of the similarity
matrix and the transportation plan. Ultimately, this leads \alg to
correctly predict that the two strings are not similar.

\begin{figure}[h]
  \begin{subfigure}[t]{0.49\textwidth}
    \centerline{\includegraphics[width=0.8\textwidth]{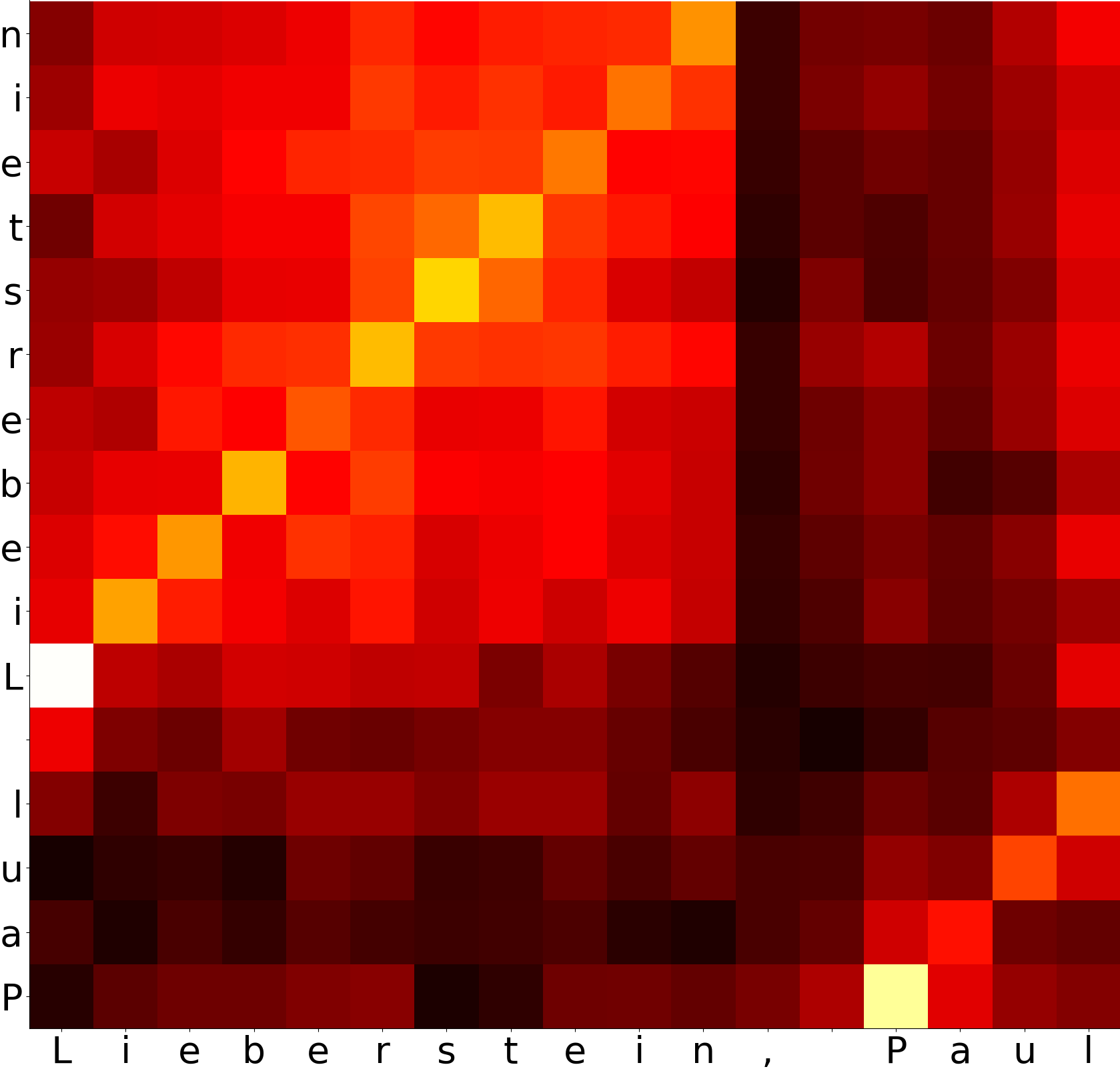}}
    \caption{Similarity}
    \label{fig:noise-filter}
  \end{subfigure}%
  \begin{subfigure}[t]{0.49\textwidth}
    \centerline{\includegraphics[width=0.8\textwidth]{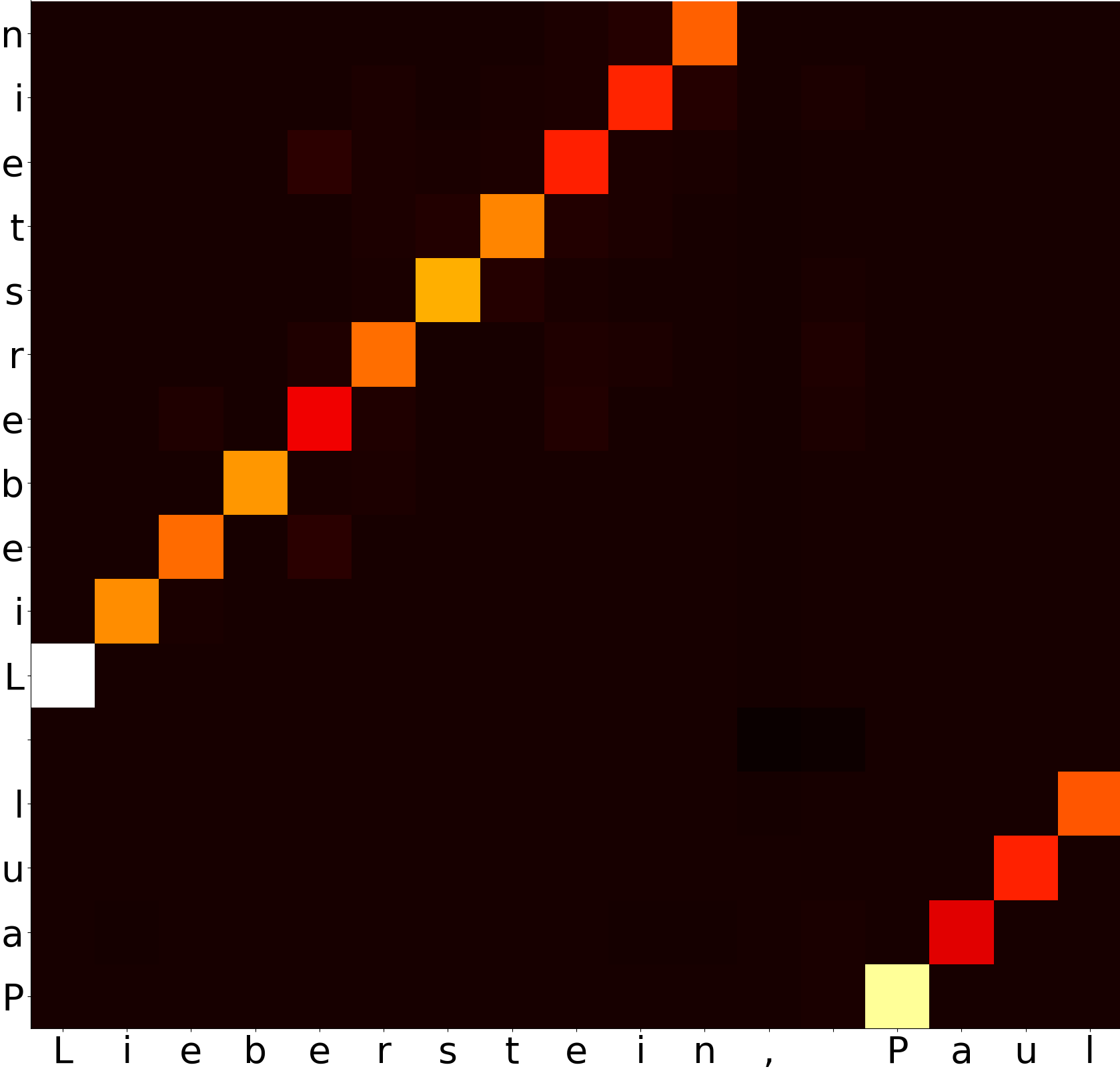}}
    \caption{Similarity x Transport}
    \label{fig:perm}
  \end{subfigure}%
  \caption{\textbf{Token Permutation:} \alg learns that token
    permutations preserve string similarity (Paul Lieberstein / Lieberstein, Paul).}
\end{figure}

\paragraph{Token Permutation.} A natural and frequently occurring
similarity-preserving edit pattern that occurs in our datasets is
token permutation, i.e., the tokens of two aliases of the same entity
are ordered differently in each mention.  For example, consider the
similarity matrix in Figure \ref{fig:perm}. The CNN easily learns that
two strings may be aliases of the same entity even if one is a token
permutation of the other. This is because it identifies multiple
contiguous ``diagonal lines'' in the similarity matrix.  Classic and
learned string similarity measures do not learn this relationship
easily.

%auto-ignore
%auto-ignore

\subsection{Cross Document Coreference}
We evaluate the impact of using \alg for in cross-document coreference
in the Twitter at the Grammy's dataset \cite{dredze2016twitter}. This
dataset consists of 4577 mentions of 273 entities in tweets published
close in time to the 2013 Grammy awards.  We use the same train/dev/test 
partition with data provided by the authors \footnote{\url{bitbucket.org/mdredze/tgx}}.
  The dataset is notable for
having significant variation in the spellings of mentions that refer
to the same entity.  We design a simple cross-document coreference
model that ignores the mention context and simply uses \alg trained on
the WikiPPL model. We perform average linkage hierarchical
agglomerative clustering using \alg scores as the linkage function and
halt agglomerations according to a threshold (i.e., no agglomerations
with linkage below the threshold are performed). We tune the threshold
on the development set by finding the value which gives the highest
evaluation score ($B^3$ F1).  We compare our method to the previously
published state of the art methods (Green \cite{green2012entity} and
Phylo \cite{andrews2014robust}). Both of these methods report numbers
using their name spelling features alone as well as with
context features. We find that our approach outperforms both methods
(including those using context features) on the test dataset in terms
of $B^3$ F1 (Table \ref{tbl:xdoc}).
%auto-ignore
\begin{table}[t]
	\footnotesize
	\centering
	\begin{tabular}{c c c}
		\bf Method & \bf Dev $B^3$ F1 & \bf Test $B^3$ F1 \\
		\hline
		Ours (HAC + \model) & 93.5 & \bf 82.5 \\
		Green (Spelling Only) & 78.0 & 77.2 \\
		Green (with Context) & 88.5 & 79.7 \\
		Phylo (Spelling Only) & 96.9 & 72.3 \\
		Phylo (with Context) & 97.4 & 72.1 \\
		Phylo (with Context \& Time) & \bf 97.7 & 72.3 \\
         \hline
	\end{tabular}
	\caption{Cross Document Coreference Results on Twitter at the Grammy's Dataset. Baseline results from \cite{dredze2016twitter}.}
	\label{tbl:xdoc}
\end{table}

%auto-ignore
\section{Related Work}
 Classic string similarity methods based on string alignment 
 include Levenshtein distance, Longest Common Subsequence,
\citet{needleman1970general}, and 
\citet{smith1981identification}. 

Sequence modeling and alignment is a widely studied problem in both
theoretical and applied computer science and is too vast to be
properly covered entirely. We note that the most relevant prior work
focuses on learned string edit models and includes the work of
\citet{mccallum2005conditional} which uses a model
based on CRFs, and \citet{bilenko2003adaptive} which uses a SVM-based
model. \citet{andrews2012name,andrews2014robust} developed a generative
model, which is used for joint cross document coreference and string
edit modeling tasks. Closely related work
also appears in the field of computational morphology
\cite{dreyer2008latent,faruqui2016morphological,rastogi2016weighting}. Much
of this work uses WFSTs with learned parameters. JRC-Names \cite{steinberger2011jrc,ehrmann2017jrc} is a dataset that stores multilingual aliases of person and organization entities.

Similar neural network architectures to our approach have been used
for related sequence alignment problems. \citet{santos2017} uses an RNN to encode toponyms before using a multi-layer perceptron to determine if a pair of toponyms are matching. The Match-SRNN computes a
similarity matrix over two sentence representations and uses an RNN
applied to the matrix in a manner akin to the classic dynamic program
for question answering and IR tasks~\cite{wan2016match}. A similar
RNN-based alignment approach was also used for phoneme recognition
\cite{graves2012sequence}. 
 Many previous works have studied
character-level models
\cite{kim2016character,sutskever2011generating}.

Alias detection also bears similarity to natural language inference tasks, where instead of aligning characters to determine if two mentions refer to the same entity, the task is to aligns words to determine if two sentences are semantically equivalent \cite{bowman2015large,williams2018broad}. 

Optimal transport and the related Wasserstein distance is studied in
mathematics, optimization, and machine learning
\cite{peyre2017computational,villani2008optimal}. It has notably been
used in the NLP community for modeling the distances between documents
\cite{kusner2015word,huang2016supervised} as the cost of transporting
embedded representations of the words in one document to the words of
the another, in point cloud-based embeddings \cite{frogner2018learning}, and in
learning word correspondences across languages and domains.
\cite{alvarez2018gromov,pmlr-v89-alvarez-melis19a}.

String similarity models are crucial to record linkage, deduplication,
and entity linking tasks. These include author coreference
\cite{levin2012citation}, record linkage in databases
\cite{li2015robust}, and record linkage systems with impactful
downstream applications \cite{sadosky2015blocking}.

%auto-ignore
\section{Conclusion}
In this work, we present \alg, a neural model of string similarity
that is trained end-to-end. The main components of our model are: a
character-level bidirectional LSTM for character encoding, a soft
alignment mechanism via optimal transport, and a powerful CNN for
scoring alignments. We evaluate our model on 5 datasets created from
publicly available knowledge bases and demonstrate that it outperforms
the baselines in almost all cases. We also show that using \alg
improves upon state of the art performance in cross-document
coreference in the Twitter at the Grammy's dataset. We analyze our
trained model and show that its optimal transport component helps to
filter noise and that is has the capacity to learn non-standard
similarity-preserving string edit patterns. 

In future work, we hope to further study the connections
between our optimal transport-based alignment method and 
methods based on attention. We also hope to 
consider connections to work on 
probabilistic latent representation of permutations and matchings \cite{mena2018learning,linderman2018reparameterizing}. Additionally, we 
hope to apply \alg to a wider-range of entity resolution tasks, for 
which string similarity is a component of model that considers additional
features such as the natural language context of the entity mention.

%auto-ignore
\subsection*{Acknowledgments} 
We thank Haw-Shiuan Chang and Luke Vilnis for their helpful discussions. 
We also thank the anonymous
reviewers for their constructive feedback.
This work was supported in part by the UMass Amherst Center for Data Science 
and the Center 
for Intelligent
Information Retrieval, in part by DARPA under agreement number
FA8750-13-2-0020, in part by Amazon Alexa Science, in part by Defense Advanced Research Agency (DARPA) contract number HR0011-15-2-0036, in part by the
National Science Foundation (NSF) grant numbers DMR-1534431 and IIS-1514053 and in part by the
Chan Zuckerberg Initiative under the project ``Scientific
Knowledge Base Construction''. The
work reported here was performed in part using high performance computing equipment obtained under a grant from
the Collaborative R\&D Fund managed by the Massachusetts
Technology Collaborative. Any opinions, findings and conclusions or recommendations expressed in this material are
those of the authors and do not necessarily reflect those of
the sponsor

\newpage

\bibliographystyle{ACM-Reference-Format}
\bibliography{references}

\newpage

\end{document}